\documentclass[conference]{IEEEtran}
\IEEEoverridecommandlockouts

\usepackage{cite}
\usepackage{amsmath,amssymb,amsfonts}
\usepackage{algorithmic}
\usepackage{graphicx}
\usepackage{textcomp}
\usepackage{xcolor}
\usepackage{hyphenat}
\usepackage{subcaption}
\usepackage{multirow}
\usepackage{lipsum}
\usepackage{booktabs,amsfonts,dcolumn}
\usepackage{tabularx}
\usepackage{makecell}
\usepackage{comment}
\usepackage{diagbox}
\usepackage{xspace}
\usepackage{gensymb}
\usepackage[group-separator={,}]{siunitx}
\usepackage[colorlinks,bookmarksopen,bookmarksnumbered,citecolor=green,urlcolor=blue]{hyperref}
\usepackage{stfloats}

\def\BibTeX{{\rm B\kern-.05em{\sc i\kern-.025em b}\kern-.08em
    T\kern-.1667em\lower.7ex\hbox{E}\kern-.125emX}}
    
\long\def\invis#1{}

\makeatletter
\DeclareRobustCommand\onedot{\futurelet\@let@token\@onedot}
\def\@onedot{\ifx\@let@token.\else.\null\fi\xspace}

\def\eg{{e.g}\onedot}
\def\ie{{i.e}\onedot}

\def\etal{\emph{et al}\onedot}

\usepackage{enumitem}

\newcommand{\iros}[1]{#1}

\DeclareUnicodeCharacter{2212}{-}
\begin{document}

\title{\LARGE NemeSys: Toward Online Underwater Exploration with\\ Remote Operator-in-the-loop Adaptive Autonomy
\thanks{\noindent\rule[4pt]{0.95\linewidth}{0.8pt}}
\thanks{This research is supported by the U.S. National Science Foundation (NSF) grants \#$2330416$ and the University of Florida Research grant \#$132763$.

$^\star$ These two authors have contributed equally.}
}


\author{\IEEEauthorblockN{Adnan Abdullah$^\star$}
\IEEEauthorblockA{\textit{Department of ECE} \\
\textit{University of Florida}\\
Gainesville, FL, USA \\
adnanabdullah@ufl.edu}
\and
\IEEEauthorblockN{Alankrit Gupta$^\star$}
\IEEEauthorblockA{\textit{Department of MAE} \\
\textit{University of Florida}\\
Gainesville, FL, USA \\
gupta.alankrit@ufl.edu}
\and
\IEEEauthorblockN{Vaishnav Ramesh}
\IEEEauthorblockA{\textit{Department of ECE} \\
\textit{University of Florida}\\
Gainesville, FL, USA \\
vaishnavramesh@ufl.edu}
\and
\IEEEauthorblockN{Shivali J. Patel}
\IEEEauthorblockA{\textit{Department of ECE} \\
\textit{University of Florida}\\
Gainesville, FL, USA \\
spatel30@ufl.edu}
\and
\IEEEauthorblockN{Md Jahidul Islam}
\IEEEauthorblockA{\textit{Department of ECE} \\
\textit{University of Florida}\\
Gainesville, FL, USA \\
jahid@ece.ufl.edu}
}

\maketitle

\begin{abstract}
Adaptive mission control and dynamic parameter reconfiguration are essential for autonomous underwater vehicles (AUVs) operating in GPS-denied, communication-limited marine environments. However, AUV platforms generally execute static, pre-programmed missions or rely on tethered connections and high-latency acoustic channels for mid-mission updates, significantly limiting their adaptability and responsiveness. In this paper, we introduce \textit{NemeSys}, a novel AUV system designed to support real-time mission reconfiguration through compact magnetoelectric (ME) signaling. We present the full system design, control architecture, and a mission encoding framework that enables interactive exploration and task adaptation via low-bandwidth communication. 
The proposed system is validated through analytical modeling, controlled simulation tests, and real-world trials. 
{The mid-mission retasking scenarios, evaluated using the NemeSys digital twin, demonstrate behavior switching latency below $50$\,ms with only a $13.2$\,MB peak computational overhead, making the framework suitable for deployment on edge computing hardware. Laboratory tank tests and open-water field trials further confirm stable control and reliable mission execution in dynamic underwater environments. These results establish the feasibility of online mission reconfiguration and highlight NemeSys as a promising step toward responsive, goal-driven adaptive underwater autonomy.}
\vspace{0.8mm}
\end{abstract}

\section{Introduction}

Autonomous Underwater Vehicles (AUVs) play a pivotal role in oceanographic research, subsea inspection, and naval operations~\cite{972073, BOVIO2006117}. Yet, existing AUV platforms operate under a limited autonomy paradigm, executing pre-programmed missions with minimal adaptation to environmental variability, unexpected events, or evolving mission objectives~\cite{abdullah2024human,xia2022virtual}. Adaptive mission control, where a human operator can interactively guide or adjust the robot's behavior during execution, offers a compelling solution to these limitations~\cite{chen2024word2wave,islam2018dynamic}. {While recent research has explored onboard decision-making and re-planning, such adaptations are typically triggered by the vehicle's local sensing and embedded reasoning policies~\cite{hernandez2017auv}. In contrast, many real-world operations require human instructions and oversight during task execution~\cite{islam2018dynamic}, as the AUV lacks contextual understanding. Examples include refining coverage during mapping or rescue missions, redirecting inspection upon detecting structural damage, or dynamically altering surveillance routines to reduce predictability~\cite{venkatesan2016auv}.}

\begin{figure}[t]
     \centering
     \includegraphics[width=\linewidth]{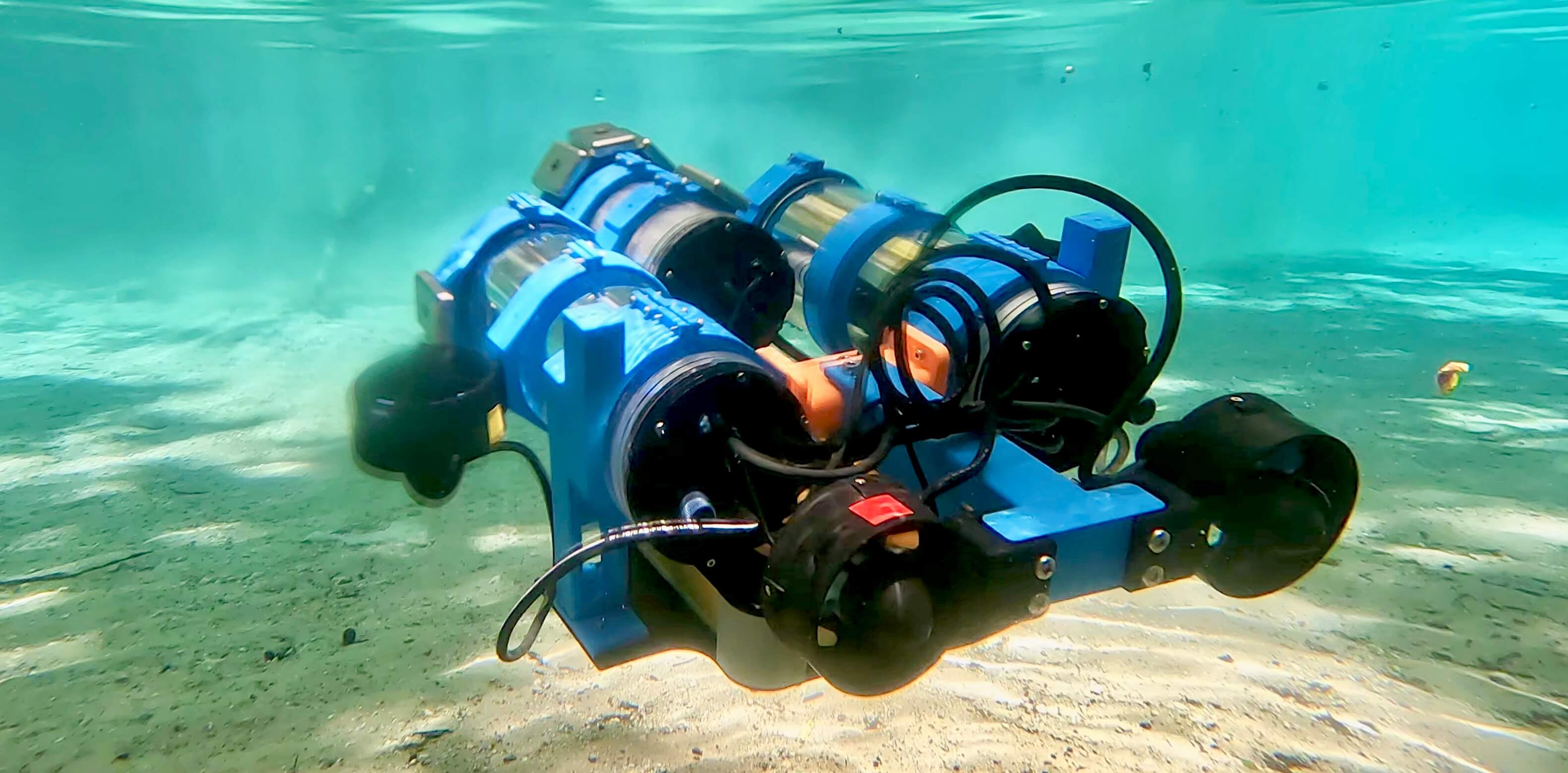}
     \includegraphics[width=\linewidth]{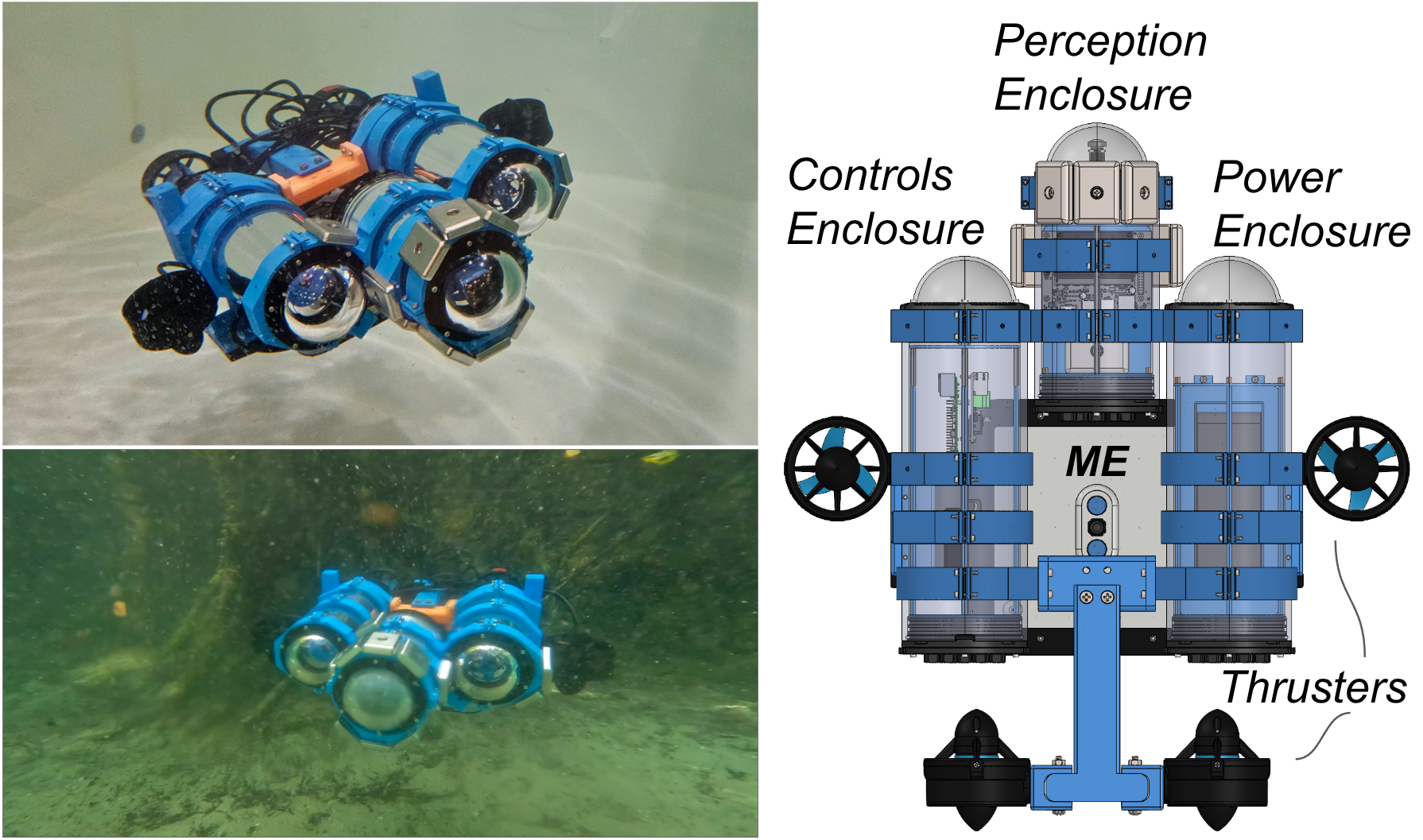}%
     \vspace{-1mm}
     \caption{Snapshots of the NemeSys AUV system deployed in the wild; major components are annotated on the right. NemeSys introduces a new class of AUVs that offers online mission-parameter updates from remote operator commands.}%
     \label{fig:fig1}
\vspace{-3mm}
\end{figure}%

Recent advances in terrestrial and aerial domains have demonstrated the advantage of integrating human inputs with autonomous system execution to improve safety and mission outcomes~\cite{rob2022}. However, the underwater domain inherently limits real-time communication as electromagnetic waves attenuate rapidly in water~\cite{10811189}. Acoustic communication offers a longer range but is constrained by low bandwidth (typically $<10$\,kbps), long propagation delays, and multipath effects~\cite{10.1145/3724121}. Optical communication can provide higher data rates but is highly sensitive to water turbidity and limited to short-range, line-of-sight conditions~\cite{wei2020led}. Consequently, existing AUVs are either constrained to fully autonomous missions with little to no adaptivity or require continuous operator oversight through tethered connections or acoustic modems, limiting their scalability and flexibility~\cite{wu2024review,zhang2022visual,islam2021robot,islam2018understanding}.

Magnetoelectric (ME) communication offers a compact and low-power alternative for transmitting structured digital commands underwater. BlueME~\cite{talebi2024blueme} demonstrates data rates of up to $36$\,kbps over $730$\,meters of range, and remains unaffected by line-of-sight or multipath issues. 
This makes it an ideal candidate for encoding compact mission updates and transmitting them during untethered AUV operation.

In this paper, we present the design and development of a new AUV platform: \textbf{NemeSys} (Fig.~\ref{fig:fig1}), designed to support untethered mission reconfiguration and adaptive task execution by low-bandwidth ME signaling. {First, we compare three candidate mechanical configurations for dynamic stability and maneuverability under integration of oil-filled ME antenna modules. The antenna introduces mass redistribution and fluid-filled inertia effects, necessitating careful evaluation of buoyancy and restoring moments~\cite{talebi2024blueme}. We then develop a lightweight mission encoding framework that represents high-level operator intents as compact digital commands suitable for low-bandwidth underwater transmission.} The operator's instruction is abstracted into a \textit{waypoint pattern} such as spiral, grid search, or perimeter scan -- and then associated with a bounded set of parameters such as speed, depth, and radius. These commands are then encoded into a binary packet using the BCH (Bose-Chaudhuri-Hocquenghem)~\cite{bose1960class} error correction scheme and transmitted over ME channel for online mission reconfiguration. 


The NemeSys platform is comprehensively evaluated, beginning with a comparative analysis across the three mechanical configurations. {We characterize inherent stability based on the AUV's response to external disturbances, followed by empirical evaluation of maneuverability through \textit{heave} and \textit{yaw} rates. We further analyze the mission-encoding framework to balance error-correction capability and coding efficiency. To validate mission adaptability at the control level, we conduct water tank experiments where the AUV performs hovering and depth transitions triggered by compact mid-mission commands. We then evaluate mission reconfiguration in Gazebo-ROS2~\cite{koenig2004design} simulation across multiple tasks, including shipwreck mapping, subsea blowout preventer (BOP) panel inspection, and surveillance of an underwater data center pod facility~\cite{abdullah2025active}. In these scenarios, we quantify the adaptive behavior in terms of command decoding latency, trajectory replanning latency, and computational overhead. Finally, we deploy NemeSys in open-water environments to validate baseline control performance and mission execution.}

\vspace{1mm}
\noindent
Overall, the main contributions are summarized as follows:
\begin{enumerate}[label={$\bullet$},nolistsep,leftmargin=*]
    \item {\textbf{Stability-aware AUV design:} We present the mechanical design of \textit{NemeSys AUV} that integrates an oil-filled ME antenna while preserving its dynamic (passive) stability. Despite the antenna-induced fluid inertia effects, NemeSys achieves a positive metacentric height and agile maneuverability: $73.1$\,mm/s heave rate and $30.3$\,deg/s yaw rate, comparable to standard compact AUVs~\cite{herbert2024design,gupta2025demonstrating}.}
    \item {\textbf{Mission encoding architecture:} We develop an encoding framework for online mission reconfiguration and parameter update over low-bandwidth ME channels. The compact $100$-bit command packet design with BCH error correction capability enables reliable, low-overhead transmission.} 
    \item {\textbf{Mission reconfiguration evaluation framework:} We establish a validation strategy to quantify online mission reconfiguration performance across mapping and inspection scenarios in digital-twin simulation, real-time depth regulation in controlled tank experiments, and long-range trajectory execution in open-water deployments.} 
    \item {\textbf{Edge-deployable adaptive autonomy:} We demonstrate less than $50$\,ms mission reconfiguration latency with a $13.2$\,MB memory footprint, validating the deployment feasibility on resource-constrained edge hardware.} 
\end{enumerate}

\vspace{1mm}
\noindent
While some existing AUV platforms, such as \textit{SUNFISH}~\cite{richmond2018sunfish}, \textit{CUREE}~\cite{girdhar2023curee}, and \textit{UX1}~\cite{martins2018ux} offer a broad range of capabilities, NemeSys offers \underline{three unique features}: (\textbf{1}) A small footprint -- one person can carry and deploy it; (\textbf{2}) Online mission reconfiguration capability during live missions, enabling adaptive task execution; and (\textbf{3}) A low-cost (below \$$3,000$), low-power design with $5$+ hours of endurance.






\begin{figure*}[t]
    \centering
    \includegraphics[width=0.95\linewidth]{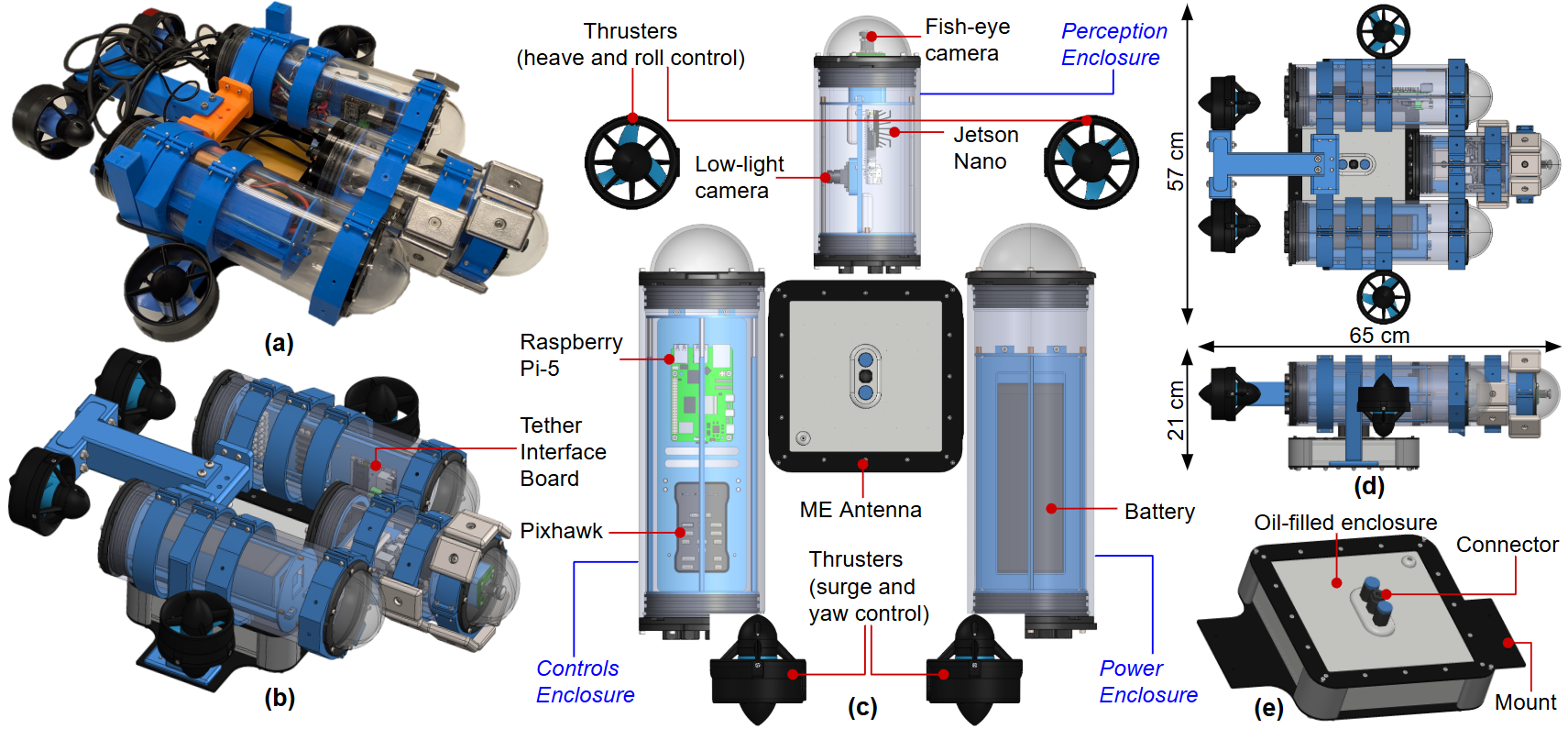}%
    \caption{System design and computational components of the NemeSys AUV: (a) The physical robot; (b) Isometric view of the corresponding CAD design showing overall structural layout; (c) Exploded view highlighting the key components; (d) Top and side views, showing the outer dimensions; and (e) ME antenna system~\cite{talebi2024blueme}.}%
    \vspace{-1mm}
    \label{fig:system_design}
\end{figure*}

\section{Related Work: AUV Systems for Autonomous Underwater Exploration}

Classical AUV architectures emphasize long-range navigation, environmental mapping, and mission execution in GPS-denied, communication-limited environments~\cite{zhou2023review}. Torpedo-style commercial platforms such as the \textit{REMUS}~\cite{allen2001remus}, \textit{Bluefin}~\cite{underwood2017design}, and \textit{NemoSens}~\cite{NemoSens} are optimized for wide-area surveys. These systems typically operate under pre-scripted waypoints with limited support for online task reconfiguration. Research platforms such as \textit{Aqua}~\cite{dudek2007aqua}, \textit{CUREE}~\cite{girdhar2023curee}, \textit{LoCO}~\cite{edge2020design}, \textit{MeCO}~\cite{widhalm2025design}, and \textit{CavePI}~\cite{gupta2025demonstrating} feature open architectures that facilitate experimentation in perception and human-robot interaction, while still lacking adaptive autonomy (mission reconfiguration) during live missions. 

\subsection{Modular and Compact AUVs}
{Compact (micro) AUVs have gained attention for operations in confined or structurally complex environments, including underwater caves, flooded mines, and subsea infrastructure. One of the earliest such platforms, \textit{UX-1}~\cite{martins2018ux}, allows for high-resolution 3D mapping of tight and maze-like underwater areas. \textit{SUAVE}~\cite{suave} emphasizes resilience and self-adaptation under system faults. The \textit{SunFish} AUV~\cite{richmond2018sunfish} is capable of collecting chemical profiles and detailed imagery inside underwater caves. \textit{Morpheus}~\cite{randeni2022morpheus}, another small-form-factor AUV, performs agile maneuvering through morphing fins and advanced control strategies.}

Other works on multi-thruster allocation and over-actuated control use quadratic programming or optimization techniques for configuration-aware control~\cite{zhu2023goa,sha2025portable}. These studies focus more on efficiency than on how the configuration and payload integration influence the motion dynamics~\cite{koshkin2023comparative,wang2021design}. The \textit{CavePI}~\cite{gupta2025demonstrating} system explores hardware and edge-AI considerations for underwater cave exploration. While it demonstrates reliable navigation and control, challenges regarding mission adaptation and dynamic reconfiguration remain open problems -- a key motivating factor of this work. 

\subsection{Online Mission Reconfiguration in AUVs}
{Adaptive mission planning for AUVs has explored onboard strategies for updating objectives in unknown environments based on observed information~\cite{albiez2010adaptive,ma2019adaptive}. Subsequent works extend this idea to autonomous re-planning for environmental monitoring, data-driven sampling, and coverage refinement~\cite{zhou2018adaptive,bai2023multi,hernandez2017auv}. For multi-AUV swarms, dynamic task assignment frameworks such as COTSBot~\cite{abbasi2020cooperative} and vision-based guidance strategies~\cite{ren2021two} allow fleets to redistribute tasks or adapt formation in response to environmental changes or vehicle failures. 
These approaches adapt mission goals by integrating model predictive control and belief-space planning to maximize coverage or minimize uncertainty in dynamic environments~\cite{zhang2022multi,mahmoudzadeh2022uninterrupted}.
} 

{Besides autonomous replanning, human-robot collaborative mission adaptation has been demonstrated in shallow-water settings, where divers adjust mission parameters through visual gesture interaction~\cite{islam2018dynamic}. Such systems rely on close-proximity interaction and do not apply for long-range operation.
In contrast, we attempt to enable operator-in-the-loop adaptive autonomy for remote missions via structured, low-bandwidth digital command transmission.}

\section{System Design}
NemeSys is a custom-designed, modular AUV featuring three enclosures as shown in Fig.~\ref{fig:system_design}. They accommodate the vehicle's perception, control, and power subsystems. 
Each enclosure consists of a $100$\,mm internal-diameter acrylic tube of thickness $6.35$\,mm sealed at both ends with aluminum end caps/acrylic domes.
The complete system weighs $13.9$\,Kg and is rated for operations at depths of up to $100$ meters.

\subsection{Perception Subsystem}
The perception subsystem (center enclosure) contains a front-facing fisheye camera, a down-facing BlueRobotics\texttrademark{} low-light camera, and an Nvidia\texttrademark{} Jetson Nano. The fisheye camera captures forward visuals with a $160^\circ$ field-of-view (FOV) and outputs a $1920\times1080$ feed at $30$\,Hz frame rate. It provides semantic understanding of the scene and detects fiducial markers (QR codes) presented by divers. The low-light camera captures downward-facing visuals with an $80^\circ \times 64^\circ$ FOV, at the same resolution and frame rate for visual feature-based state estimation. The Jetson Nano executes all image processing tasks for scene perception and localization.

\subsection{Controls and Power Subsystem}
The control subsystem (left enclosure) contains a Raspberry Pi5, a Pixhawk\texttrademark{} flight controller, a BlueRobotics\texttrademark{} tether interface board (TIB), electronic speed controllers (ESCs), a Bar30 pressure sensor, and voltage regulators. The Pi5 executes the vehicle's planning and control algorithms for real-time underwater navigation. The Pixhawk flight controller serves as the hardware-software interface: it receives actuation commands from the Pi5 via the {\tt MAVLink} communication protocol and drives the thrusters accordingly. It also integrates a $9$-DOF IMU to compute attitude. The Bar30 sensor delivers pressure measurements with $0.2$\,mbar resolution and $\pm2$\,mm accuracy. The TIB establishes a bidirectional tethered link for optional use as an ROV. 

The third enclosure houses a $14.8$\,V, $10$\,Ah BlueRobotics\texttrademark{} battery pack. Onboard voltage regulators step down this supply to power both internal systems (cameras, computers) and external loads (thrusters). At full capacity, NemeSys delivers over five hours of endurance as shown in Table~\ref{tab:battery_power_consumption_table}. Although the battery can support longer operation, we limit discharge to above $20$\% capacity as per the manufacturer's recommendations~\cite{BR_battery} to avoid rapid depletion.

\begin{table}[h]
\centering
\caption{
The battery power consumption characteristics of NemeSys (at maximum capacity) over time.}%
\vspace{-1mm}
\begin{tabular}{c|ccccccc}
\toprule
\textbf{Time (hours)} & 1 & 2 & 3 & 4 & 5 \\
\midrule
\textbf{Battery \%} & 71 & 54 & 42 & 32 & 22 \\
\bottomrule
\end{tabular}
\label{tab:battery_power_consumption_table}
\end{table}

\subsection{Communication \& Locomotion Subsystem}
The envisioned wireless communication between NemeSys and external robotic agents or operators is facilitated by BlueME~\cite{talebi2024blueme}, a low-power antenna designed for very-low-frequency (VLF) magnetoelectric communication in water medium. The antenna module is placed beneath the robot, aligned with the center of gravity to ensure hydrodynamic stability; see Fig.~\ref{fig:system_design}\,(d). Meanwhile, the structural gap among the three enclosures provides an unobstructed line of sight to the water surface, thereby enhancing the communication reliability. The antenna receives high-level mission directives as digital symbols (bitmaps), which are then decoded onboard into actionable parameters for downstream tasks.  

\begin{figure}[t]
     \centering
     \includegraphics[width=\linewidth]{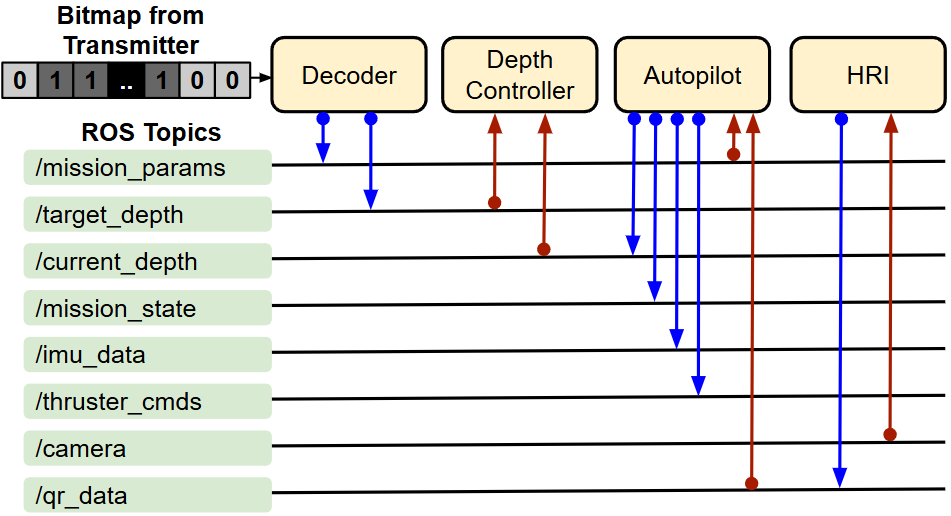}%
     \caption{Data flow among the perception, communication, and control subsystems of NemeSys is illustrated by ROS \texttt{nodes} and \texttt{topics}; red and blue arrows represent \textit{subscribed} and \textit{published} topics in the ROS graph, respectively. 
     }%
     \vspace{-1mm}
     \label{fig:ros_graph}
 \end{figure}

For autonomous mission execution, the communication, locomotion, and control subsystems are integrated in a ROS2 middleware. This architecture ensures online synchronization among the operator's input, sensor feedback, control commands, and mission state updates. An overview of the ROS node-topic architecture for autonomous operations is illustrated in Fig.~\ref{fig:ros_graph}. Upon receiving the transmitted bitmaps, a decoder node translates them into mission parameter topics, which are subscribed to by the autopilot node. The autopilot node handles low-level control such as maintaining depth and ensuring roll/pitch stability, and concurrently publishes sensory data. The final design of NemeSys comprises four thrusters that provide four degrees of freedom for locomotion: surge, heave, yaw, and roll. This low-level control is managed by Pixhawk\texttrademark{} flight controller. Additionally, a Human-Robot Interaction (HRI) node manages visual communication with divers through fiducial markers, including QR codes and ArUco tags~\cite{garrido2014automatic}.



\section{Stability-aware Design Evaluation}
\label{sec: dynamic_performance}
\iros{We characterize the motion dynamics of NemeSys by focusing on two key aspects: stability and maneuverability. Integration of the oil-filled ME antenna introduces additional mass, surface area, and fluid displacement, which alter the AUV's center of gravity and compromise stability. Hence, we analyze these motion dynamics of three candidate configurations and present their quantitative comparison.}

\begin{figure*}[t]
     \centering
     \includegraphics[width=0.95\linewidth]{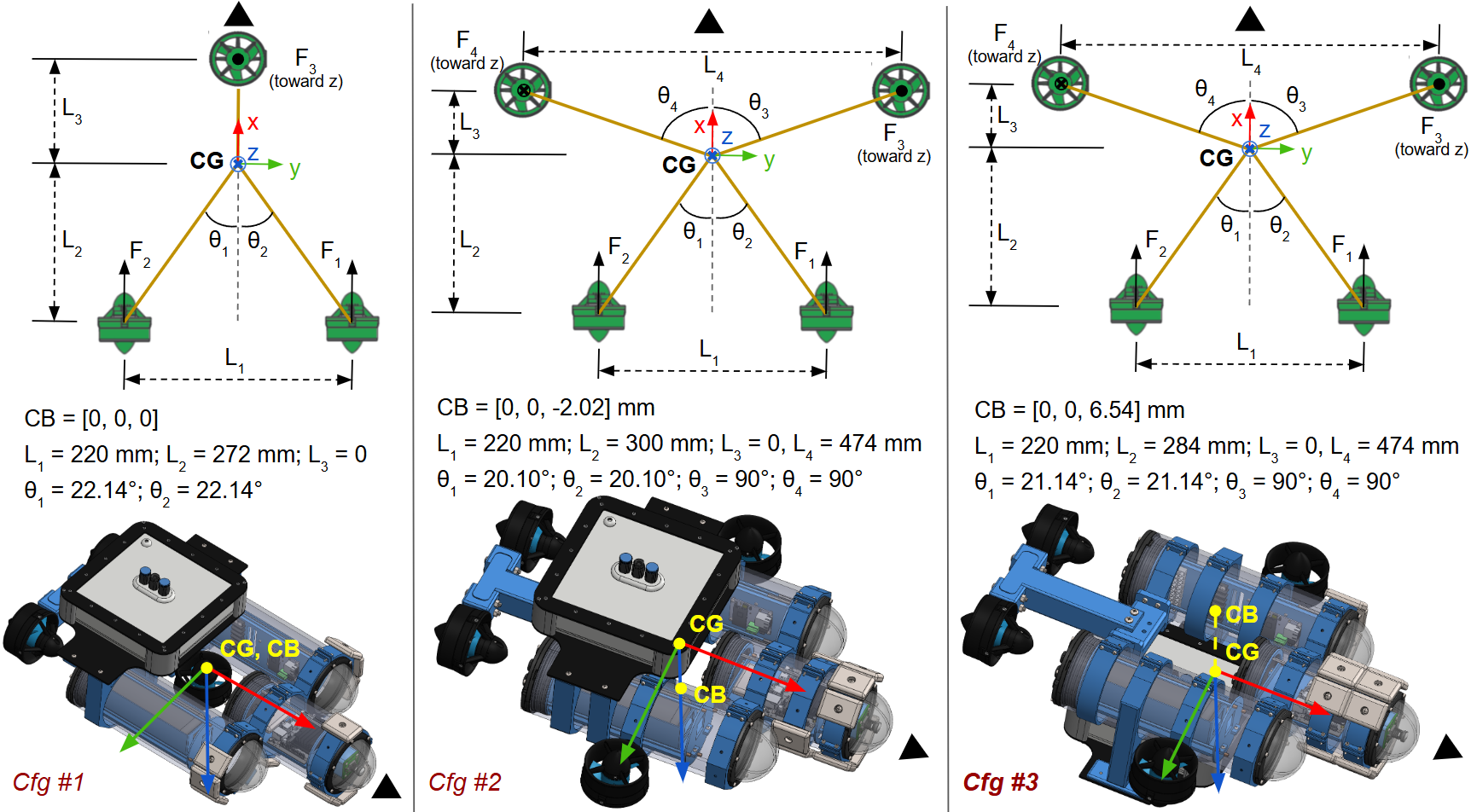}%
     \caption{The three design architectures are compared for passive stability and maneuverability. The configurations illustrate the actuator layout, thrust directions, and the relative positioning of the center of gravity $CG$ and center of buoyancy $CB$. Configuration \#3 exhibits the highest stability due to its lower CG and high restoring moment.
     }%
     \vspace{-2mm}
     \label{fig:design_comparison}
 \end{figure*}

\subsection{Dynamic (Passive) Stability}\label{subsec:stability}
The dynamic stability of an AUV determines its ability to return to equilibrium without control actuation after being subjected to small disturbances. The stability depends on the relative locations of its center of gravity (CG) and center of buoyancy (CB). Thruster placement, battery ballast, and hull geometry shape the vehicle's mass distribution and displaced‑volume centroid; consequently, both CG and (to a lesser extent) CB can vary with the mechanical layout~\cite{9977030}.

We consider NemeSys AUV in quiescent water, with weight $W = mg$ and buoyant force $B = \rho V g$, where $V$ is the displaced water volume. The vertical separation between CB and CG, $h = z_B - z_G$, is considered positive when CG lies \emph{below} CB. For a small rotation $\theta$ about any horizontal body axis (roll or pitch), the lines of action of $W$ and $B$ are displaced laterally by $h\sin\theta$. The resulting restoring moment is given by:
\begin{equation}
M = W h \sin\theta \approx W h \theta, \quad |\theta| \ll 1.
\label{eq: restoring_moment_eq}
\end{equation}
This formulation is the submerged analogue of the classical ship‑stability formula $M = W \cdot GM \cdot \theta$; here the metacentric height is simply the fixed offset $h$, as CB shifts negligibly for small angles in a fully submerged rigid body~\cite{LEONARD1997331}; Fig.~\ref{fig:design_comparison} illustrates the relative position of CB and CG across three candidate configurations.


\subsection{Controllability}
Controllability refers to the system's ability to generate independent actuator forces and moments across various DOF. A lower degree of controllability limits the AUV's capacity to reject disturbances and perform complex maneuvers. The controllability of NemeSys is evaluated following the method proposed by Deng~\etal~\cite{deng2025thrust}, which involves constructing a thruster configuration matrix $B$ that captures the contribution of each thruster to force and moment generation:    
\begin{equation}
B = \begin{bmatrix}
    f_1 & f_2 & ... \\
    r_1 \times f_1 & r_2 \times f_2 & ... 
\end{bmatrix},
\label{eq: controllability_matrix}
\end{equation}
where $f_i$ denotes the unit thrust vector of the $i^{th}$ thruster, and $r_i$ represents its position vector relative to the vehicle's center of gravity. The rank of matrix $B$ indicates the number of independently controllable DOFs. A full rank of $6$ implies complete controllability in all six DOFs. For NemeSys, the configuration achieves a rank of $4$, enabling effective control in surge, heave, roll, and yaw. This design avoids actuator redundancy while ensuring sufficient maneuverability for the intended operations.

\subsection{Maneuverability}\label{subsec:maneuverability}
\iros{Maneuverability, defined as the ability to execute rapid heading and depth adjustments in response to control commands, is critical for agile operation in dynamic underwater environments. Enhanced yaw agility improves path-following accuracy and accelerates convergence toward target headings, while robust vertical agility enables swift depth transitions for obstacle avoidance and precise station‑keeping. Randeni et al.~\cite{randeni2022morpheus} proposed calculating the yaw rate $\omega$ as a quantifiable measure of the lateral maneuverability (see~\ref{eq:maneuverability}); here $\Delta \psi$ is the change in heading angle (from IMU logs), and $\Delta t$ is the time interval over which the turn occurs. 
\begin{equation}
\omega = \frac{\Delta \psi}{\Delta t},
\qquad
\dot z = \frac{\Delta z}{\Delta t}.
\label{eq:maneuverability}
\end{equation}
Similarly, the vertical maneuverability is characterized by the heave rate $\dot z$ (see~\ref{eq:maneuverability}), where $\Delta z$ is the change in depth (from pressure sensor logs), and $\Delta t$ is the elapsed time for that depth change. Higher values indicate more responsive depth control and faster vertical maneuvers.}
 

\begin{figure}[t]
     \centering
     \includegraphics[width=\linewidth]{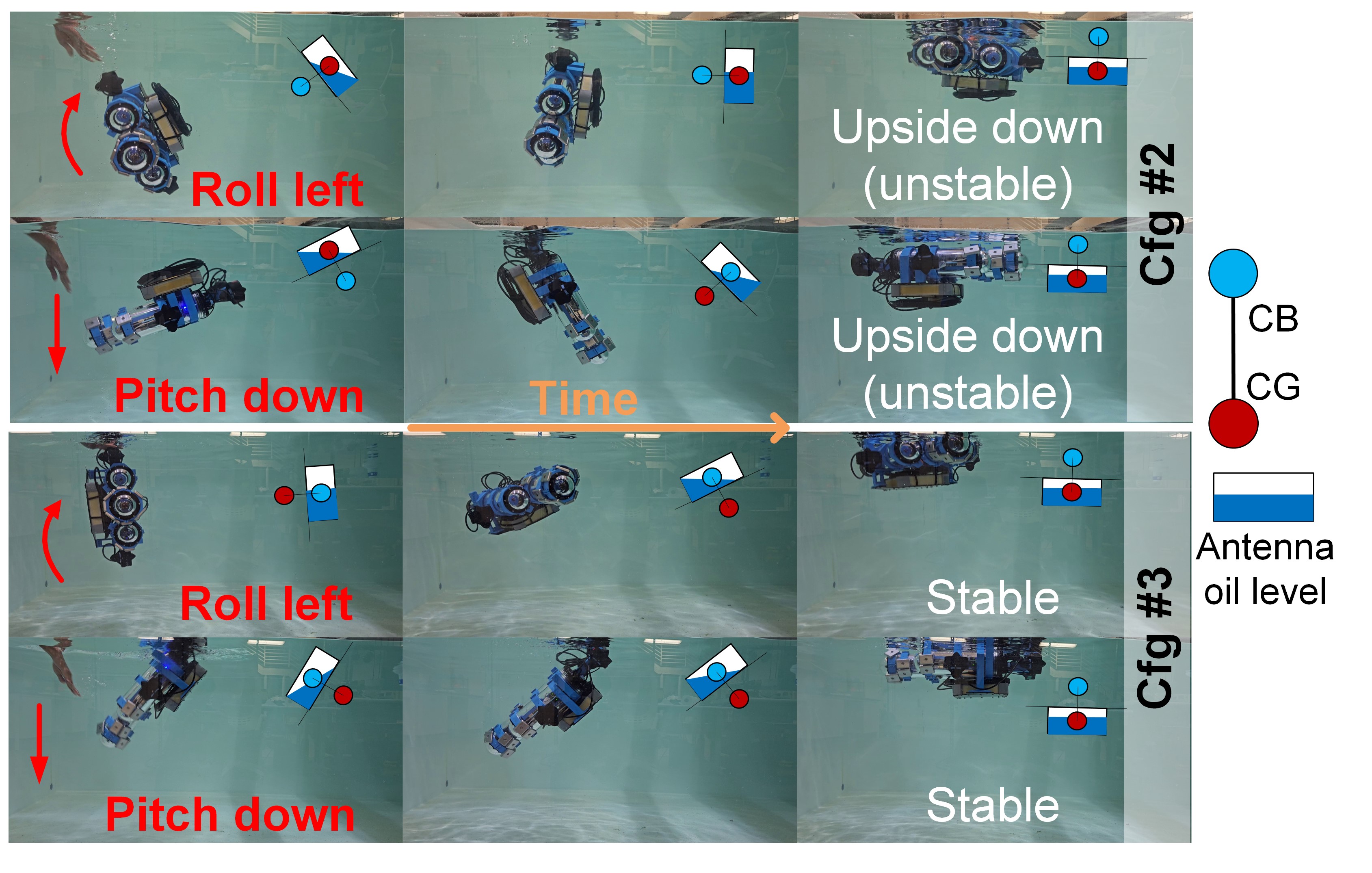}%
     \vspace{-2mm}
     \caption{Passive stability comparison under external disturbance: Cfg \#2 flips upside down, while Cfg \#3 maintains posture due to a more favorable CG and CB distribution.
     }%
     \vspace{-3mm}
     \label{fig:stability_comparison}
 \end{figure}

\subsection{Stability and Maneuverability Evaluation}\label{subsec:design_evaluation}
The NemeSys platform is developed through iterative analytical and experimental evaluation of three mechanical configurations, as summarized in Fig.~\ref{fig:design_comparison}. The first configuration employed three thrusters: a surge-yaw pair and a single heave thruster. Although the CG and CB were vertically aligned, the placement of the ME antenna above the heave thruster caused flow interference, resulting in limited heave performance; see Table~\ref{tab:design_comparison}.

\begin{table}[h]
\centering
\caption{Comparison of the three NemeSys design configurations shown in Fig.~\ref{fig:design_comparison}. Configuration \#3 demonstrates superiority across all metrics.
}
\label{tab:design_comparison}
\footnotesize
\begin{tabular}{c|c|c|c}
\hline
\textbf{Parameter} & Cfg \#1 & Cfg \#2 & Cfg \#3 (Final) \\
\hline
\textbf{Metacentric height (mm) ($\uparrow$)}      & 0 & -2.0 & 6.5 \\
\textbf{Heave rate (mm/s) ($\uparrow$)}      & 3.5 & 26.3 & 73.1 \\
\textbf{Yaw rate (deg/s) ($\uparrow$)}      & 5.7 & 9.9 & 30.3 \\
\hline
\end{tabular}
\vspace{-2mm}
\end{table}

The second iteration introduced a four-thruster layout with two vertical thrusters, eliminating the flow interference problem. However, positioning the ME antenna above the hull shifted the CG above the CB, resulting in a negative metacentric height and passive instability as shown in Fig.~\ref{fig:stability_comparison}. 

The final design resolves these issues by repositioning the ME antenna beneath the hull, which lowers the CG below the CB. This configuration results in a positive metacentric height, improving hydrostatic stability while offering control over four degrees of freedom.

Table~\ref{tab:design_comparison} summarizes the three configurations, highlighting the progressive improvements in dynamic stability (metacentric height) and maneuverability (heave rate and yaw rate). We measure heave rate by commanding a $1.5$\,m descent at full thrust, while yaw rate is obtained from maximum-thrust rotation at $0.5$\,m depth over $60$\,s.

\begin{figure*}[t]
    \centering
    \includegraphics[width=\linewidth]{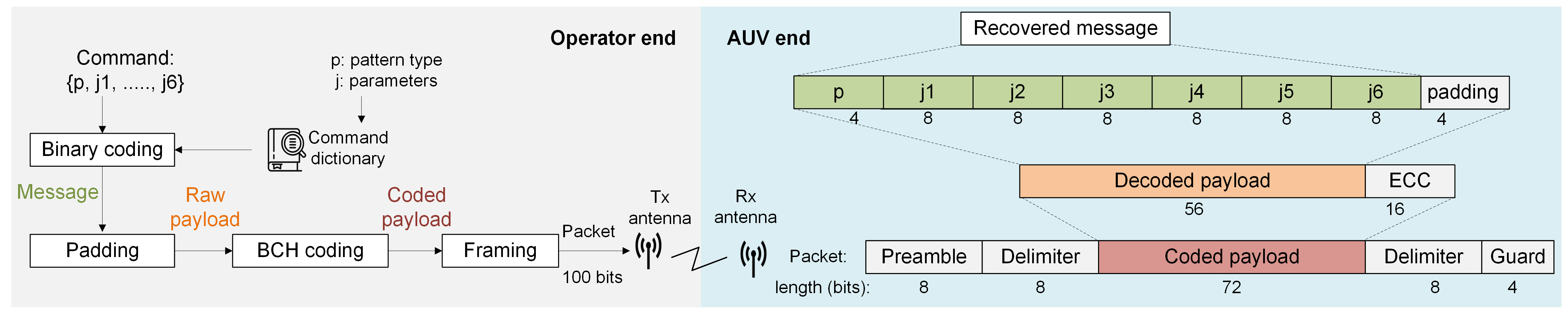}%
    \vspace{-1mm}
    \caption{The mission encoding and decoding scheme for ME-based adaptive autonomy is shown. Pattern types and associated parameters, issued by the operator, are coded into a bitstream and transmitted via the ME antenna. The receiver ME antenna onboard NemeSys decodes the instruction and adjusts its mission goals and subtasks.}%
    \label{fig:coding_scheme}
    \vspace{-1mm}
\end{figure*}

\begin{table*}[b]
\centering
\caption{Each pattern type in the \textbf{command dictionary} is assigned a unique 4-bit ID and is parameterized by up to six values. The 4-bit encoding allows for up to $16$ distinct pattern types; $8$ representative examples are presented here.}
\label{tab:command_dict}
\begin{tabular}{c||c|c|c|c|c|c|c|c}
\hline
\textbf{Pattern Type} & straight & square & lawnmower & circle & spiral & helix & hover & box orbit \\
\textbf{Pattern ID}   & 0000 & 0001 & 0010 & 0011 & 0100 & 0101 & 0110 & 0111 \\
\hline
\hline
\textbf{$j_1$}      & cruise speed & cruise speed & cruise speed & cruise speed & cruise speed & cruise speed & duration & cruise speed \\
\textbf{$j_2$}      & target depth & target depth & target depth & target depth & target depth & start depth & target depth & target depth \\
\textbf{$j_3$}      & duration & side span & sweep length & radius & initial radius & end depth & heading & radius \\
\textbf{$j_4$}      & heading & direction & lane spacing & direction & final radius & radius & N/A & direction \\
\textbf{$j_5$}      & N/A & N/A & \# laps & N/A & \# loops & turns & N/A & \# laps \\
\textbf{$j_6$}      & N/A & N/A & N/A & N/A & direction & direction & N/A & N/A \\
\hline
\end{tabular}
\end{table*}

\section{Adaptive Mission Configuration \& Analyses} \label{sec:mission_reconfig}

\iros{This section presents the mission encoding framework that maps operator intent into digital commands for mid-mission updates.} The operator's instructions (\eg, mission modes and control parameters) are transmitted via the ME antenna from a surface station or buoy, decoded onboard, and executed in real time to update mission goals.

\subsection{Pattern Encoding}

The encoder is designed to address two primary constraints of the underwater ME channel: data rate and operational range; the setup is limited to $36$\,kb/s and $730$\,m, respectively~\cite{talebi2024blueme}. To ensure timely response during mission-critical operations, we impose an upper limit of $1$\,ms per command transmission, which aligns with latency requirements for mid-range underwater communication systems~\cite{stojanovic2009underwater,basagni2019marlin}.
 

As illustrated in Fig.~\ref{fig:coding_scheme}, each command is structured into a pattern type $\tt{p}$ and associated parameters $\tt{j}$. The pattern types include but are not limited to square, circle, grid-search (lawnmower), spiral, and helix (see Table~\ref{tab:command_dict}). 
Each command begins with a $4$-bit identifier for the pattern type, followed by up to six $8$-bit parameter fields, forming a raw payload of $52$ bits (zero-padded to match $56$ bit BCH message length). This payload is encoded using forward error correction (FEC)~\cite{lin2001error} to produce a $72$-bit codeword. Finally, the codeword is framed with preamble, delimiter, and guard bits, resulting in a complete $100$-bit transmission packet. This low-overhead encoding scheme achieves reliable communication with minimal packet loss. The use of FEC further ensures resiliency in the presence of signal degradation, which is common in underwater propagation environments~\cite{van2013propagation,lanzagorta2012underwater}.

\subsection{Error Correction \& Online Pattern Decoding}\label{subsec:ecc}
We adopt a binary BCH FEC scheme~\cite{bose1960class} for error correction in the presence of signal degradation. BCH codes are a subclass of cyclic error-correcting codes capable of correcting multiple random bit errors~\cite{chien2003cyclic}. 

Specifically, a $(n=72, k=56, T=2)$ is used, where $k$ denotes the payload length, $n$ the codeword length, and $T$ the maximum number of correctable bit errors. This encoding process appends $n-k = 16$ redundancy bits to the message polynomial $\mathbf{m}(x)$ of length $k$, producing a codeword $\mathbf{c}(x)$ of length $n$ such that $\mathbf{c}(x) \mod g(x) = 0$, where $g(x)$ is the generator polynomial (\ref{eq:bch_encoding}).
\begin{equation}
\mathbf{c}(x) = \mathbf{m}(x) \cdot x^r + \texttt{modulus} \big( {\mathbf{m}(x) \cdot x^r},~{g(x)} \big)
\label{eq:bch_encoding}
\end{equation}
At the receiver, a possibly corrupted polynomial $\mathbf{r}(x) = \mathbf{c}(x) + \mathbf{e}(x)$ is used to compute the syndromes $S_i$ using:
\begin{equation}
S_i = \mathbf{r}(\alpha^i), \quad i = 1, 2
\label{eq:syndrome}
\end{equation}
where $\alpha$ is a primitive element of the Galois field $\text{GF} (2^m)$~\cite{lin2001error}. These syndromes are then used in the Berlekamp–Massey algorithm~\cite{massey2003shift} to compute the error locator polynomial, followed by the Chien search~\cite{chien2003cyclic} to identify and correct the error locations. This process allows recovery of the original message with up to two bit errors per codeword.


\subsection{Encoding Scheme Analyses} 
\begin{figure}[t]
     \centering
     \includegraphics[width=\linewidth]{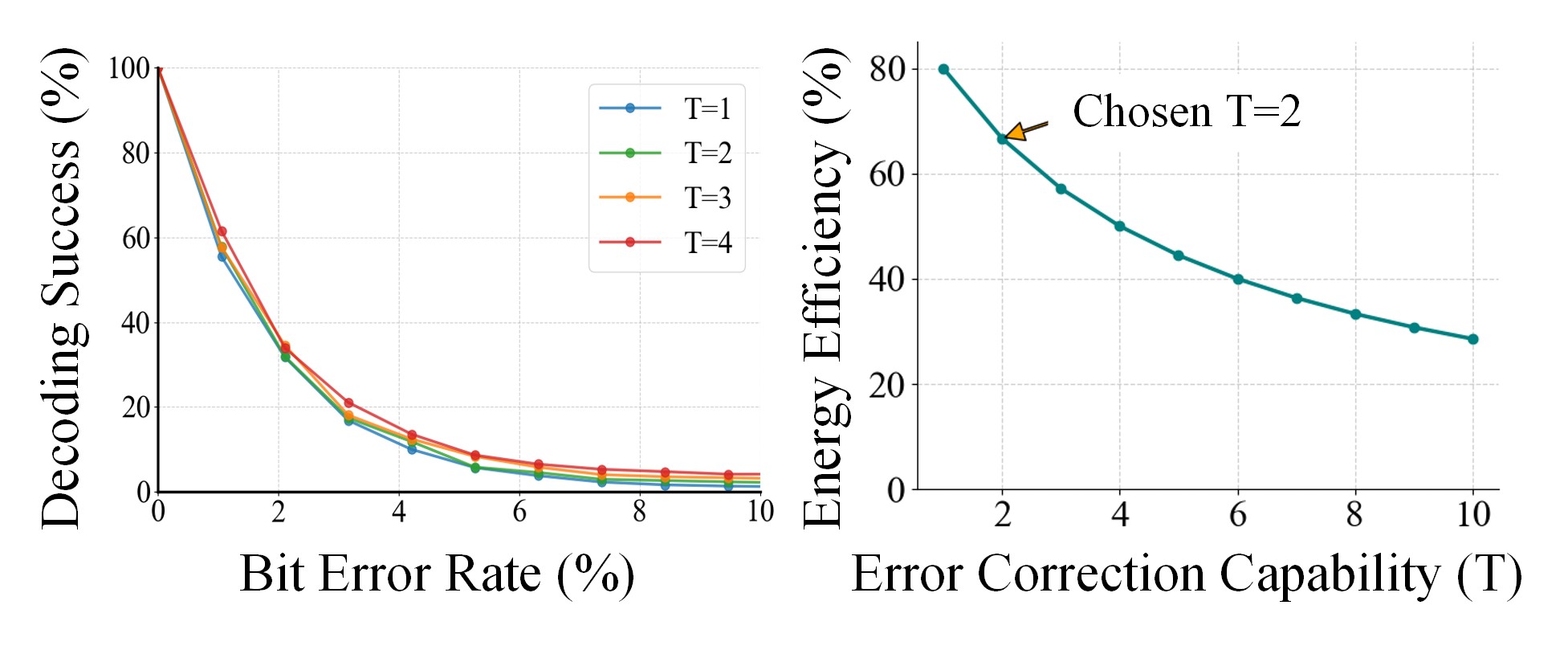}%
     \caption{The choice of error correction strength $T=2$ offers a favorable trade-off: it improves decoding success over $T=1$ at more commonly encountered low BER region, with higher efficiency compared to stronger codes with $T>2$.
     }%
     \vspace{-1mm}
     \label{fig:encoding_eval}
 \end{figure}

The encoding scheme is designed to meet the bandwidth constraint of the ME channel while maintaining reliable decoding under noisy conditions. 
Fig.~\ref{fig:encoding_eval} shows the decoding success rate under varying bit error rates (BER). While stronger error correction methods slightly improve success rates, they reduce efficiency due to added redundancy. The coding efficiency is defined as $k / (k + 2mT)$; where $m$ is $7$ for $56$-bit payload. 
Therefore, we choose $T = 2$ as a balanced configuration that maintains efficiency over $60$\%.

\section{End-to-End System Validation}

To validate the proposed mission reconfiguration framework, we conduct multi-tier experiments spanning Gazebo-ROS2 simulation, controlled water-tank testing, and open-water deployment. This structured pipeline assesses \textbf{(i)} mid-mission reconfiguration latency and computational overhead with NemeSys digital twin, \textbf{(ii)} real-time parameter update performance in water tank, and (\textbf{iii}) open-water trajectory execution under real environmental disturbances.

\begin{figure*}[t]
     \centering
     \includegraphics[width=\linewidth]{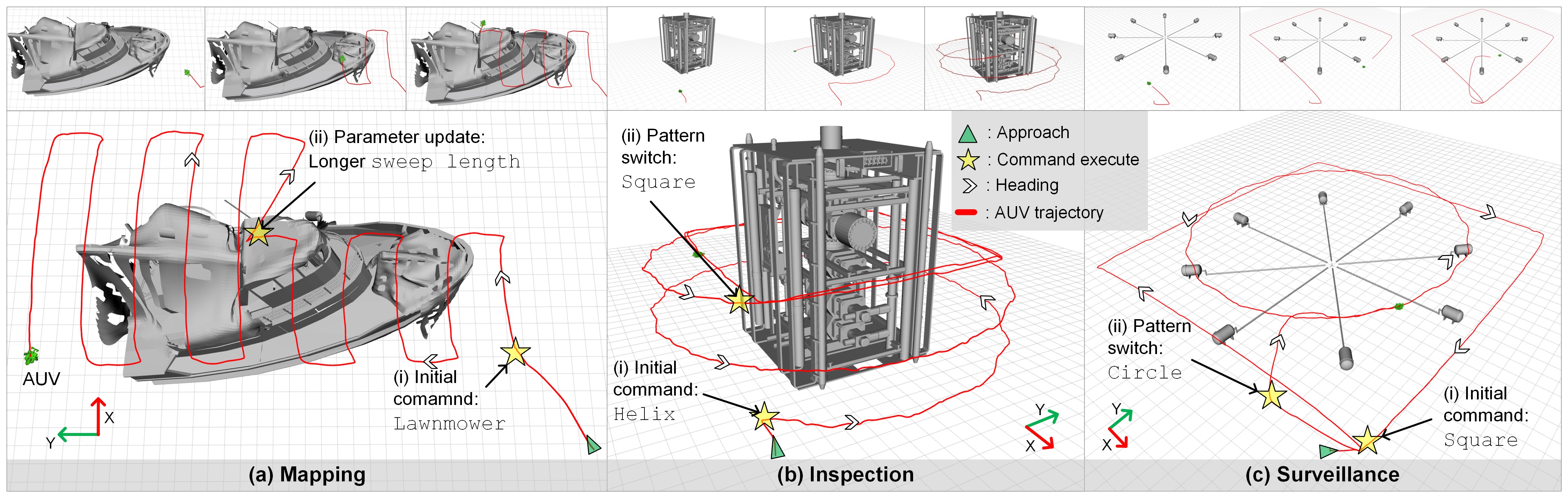}%
     \caption{\iros{Online mission reconfiguration is demonstrated in three representative scenarios with the NemeSys digital twin: (a) expansion of shipwreck mapping via an increased \texttt{lawnmower sweep length}; (b) transition from a 3D \texttt{helical} inspection of a BOP panel to a constant-depth \texttt{square} pattern for a close-up look; and (c) transition from a \texttt{square} patrol around a subsea pod facility to a \texttt{circular} pattern at a different depth to reduce predictability. Initial and mid-mission commands for each scenario are denoted by (i) and (ii), respectively; intermediate snapshots are shown on the top row.}
     }%
     \vspace{-1mm}
     \label{fig:sim_reconfig}
 \end{figure*}

\subsection{Reconfiguration Delay Analyses with Digital Twin}
\iros{We conduct controlled experiments with the Gazebo-based digital twin of NemeSys to benchmark the reconfiguration performance. The twin reflects the mechanical properties of the physical system, including mass, buoyancy, and thrust forces. 
The onboard sensor suite -- comprising a front-facing camera, downward camera, IMU, and pressure/depth sensor -- is emulated using Gazebo's native sensor plugins. 
A $60$\,m deep open water environment with two different structures (shipwreck and BOP panel) is sourced from the UUV simulator~\cite{manhaes2016uuv}.} 

\iros{In these experiments, the received codeword from the ME channel is emulated by publishing a ROS topic, thereby isolating the mission adaptation pipeline from the physical communication layer. Fig.~\ref{fig:sim_reconfig} presents three representative mid-mission retasking scenarios: \textbf{(a)} parameter-level update: lawnmower sweep length extension for mapping; \textbf{(b)} pattern-level switch: 3D helix to square inspection; and \textbf{(c)} combined pattern and depth update: square to circle surveillance.} 


\iros{To quantify the mission reconfiguration performance, we define and measure the following metrics:
\begin{itemize}[nolistsep, leftmargin=*]
    \item \textbf{Decode latency} is the time elapsed from command reception to a new mission acknowledgment. It captures the time required for bitarray conversion, BCH error correction, and payload parsing. Since all commands use fixed-length codewords, we report the average over $30$ trials.
    \item \textbf{Replan latency} is the time elapsed from mission acknowledgment to the first updated control output. It scales with waypoint count and planning complexity. For fair comparison, we report a \textit{normalized replan latency}, defined as the total replanning time divided by the number of waypoints.
    \item \textbf{Memory requirement} is quantified by the peak resident set size (RSS) during the decoding and replanning process. It captures the physical memory footprint of the software stack, including the decoder and waypoint generator.
\end{itemize}
}

\begin{table}[h]
\centering
\caption{\iros{Mission adaptation performance in simulation: replan latency is averaged over $10$ trials per scenario; decode latency and memory are scenario independent, therefore averaged over all trials. The tests are performed on Ubuntu~$20.04$ with an Intel\textsuperscript{\textregistered} Core\textsuperscript{\texttrademark} i7 CPU and $16$\,GB RAM.}}
\label{tab:reconfig_metrics}
\footnotesize
\begin{tabular}{c|c|c|c}
\hline
\textbf{Scenario} & 
\textbf{Normalized Replan} & 
\textbf{Decode} & 
\textbf{Memory} \\
 & 
\textbf{Latency (ms)} & 
\textbf{Latency (ms)} & 
\textbf{(MB)} \\
\hline
Shipwreck 
& 42 
& \multirow{3}{*}{0.027} 
& \multirow{3}{*}{13.2} \\
\cline{1-2}

BOP 
& 15 
&  
&   \\
\cline{1-2}

Pod 
& 46 
&  
&   \\
\hline
\end{tabular}
\vspace{-1mm}
\end{table}

\vspace{1mm}
\noindent
\textbf{Observations}. The quantitative results are summarized in Table~\ref{tab:reconfig_metrics}. These experiments reveal some key findings about the proposed mission reconfiguration framework. Since all commands are encoded using fixed-length 100-bit packets, the decoding time remains consistent ($0.027$\,ms) across all scenarios and command types.  Unlike decoding, replanning time depends on the number of generated waypoints. It exhibits non-linear growth with trajectory complexity. Therefore, we observe variability in normalized replanning latency. For instance, a $4$-waypoint square pattern yields a replan latency of as low as $26$\,ms (\ie, $6.5$\,ms normalized latency),  whereas a $36$-waypoint circular pattern may incur a latency of up to $1.62$\,s (\ie, $45$\,ms normalized latency).

Across all scenarios, the system achieves less than $50$\,ms total latency. The peak memory usage during decoding and replanning remains limited to $13.2$\,MB, indicating suitability for edge device deployment. These experiments demonstrate that operator intent, encoded as compact bitmaps, can be used to adapt the AUV's mission online without requiring high-bandwidth streaming or mission restarts.

\begin{figure}[t]
     \centering
     \includegraphics[width=\linewidth]{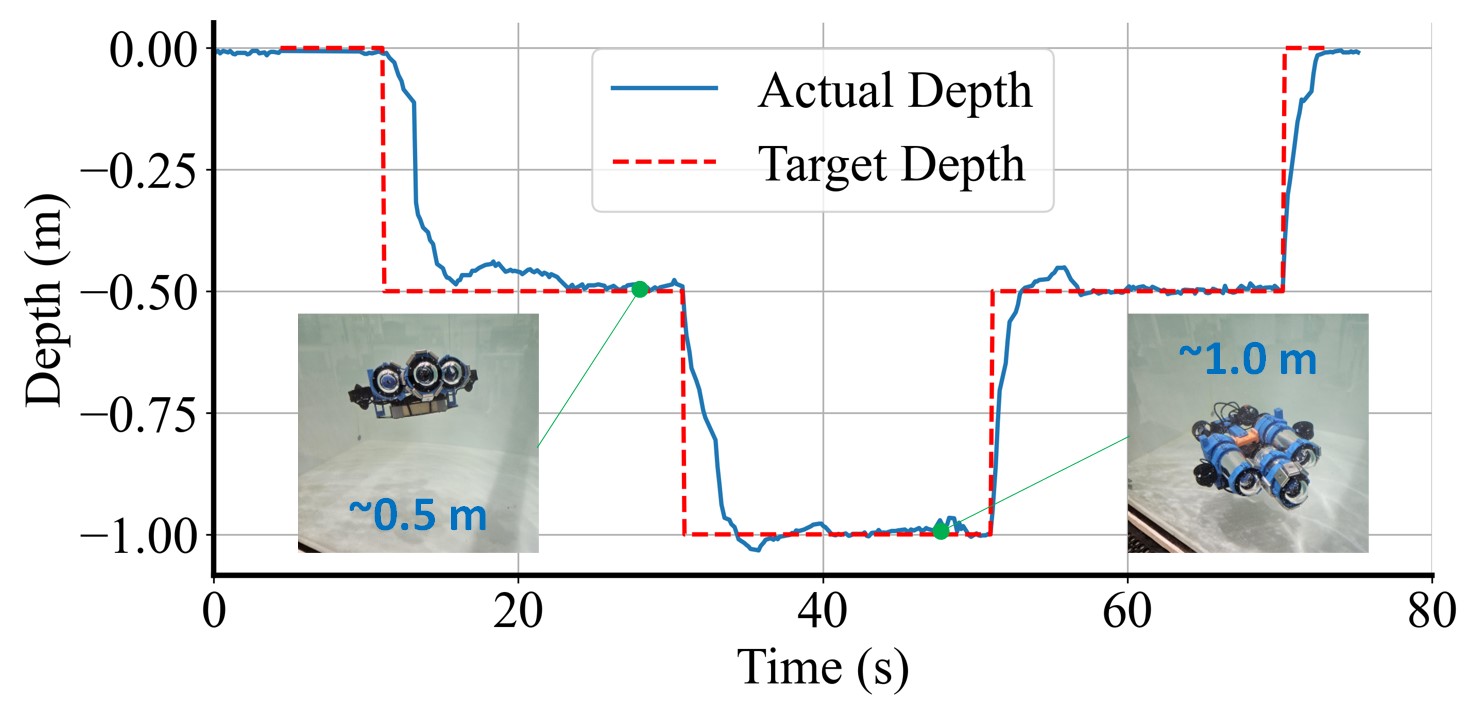}%
     \vspace{-1mm}
     \caption{\iros{The AUV's response to a sequence of target depth update commands during hovering is tested over a $75$-second window. The controller tracks target depths of $0.5$\,m and $1.0$\,m with transient periods of approximately $5$ seconds and maintains steady-state deviations within $\pm3$\,cm.}
     }%
     \vspace{-2mm}
     \label{fig:depth_control}
 \end{figure}
 
\subsection{Bench Test: Adaptive Control}

The mission parameter update capability is tested in a $3$\,m$\times2$\,m$\times1.5$\,m laboratory water tank. Given the small tank dimensions and positively buoyant nature of NemeSys, we fix the motion pattern to simple \texttt{hovering} and switch the target depth in real-time to evaluate mission-level adaptivity. Depth regulation is achieved using two laterally mounted thrusters that generate the required heave force, with a proportional-derivative controller providing closed-loop depth stabilization.

\iros{Each command is encoded as a fixed-length binary message of the form $\{\tt{p, j_1, \ldots, j_6}\}$, as described in Table~\ref{tab:command_dict}. For this experiment, the target depth corresponds to the second parameter ($j_2$), which is modified at four time instances by the operator. Fig.~\ref{fig:depth_control} illustrates the $75$\, seconds timeline of the commands sent and the resulting depth response. The controller tracks a sequence of step changes in the commanded depth, between $0$ to $1.0$\,m range. The results demonstrate smooth transitions and stable depth regulation under repeated parameter updates, validating the proposed mission parameter update scheme. Note that the actual ME transmission channel is not utilized in this experiment. Instead, the mission commands are encoded and published as ROS topics and are subscribed to by the AUV's onboard computer for execution.}


\begin{figure}[t]
     \centering
     \includegraphics[width=\linewidth]{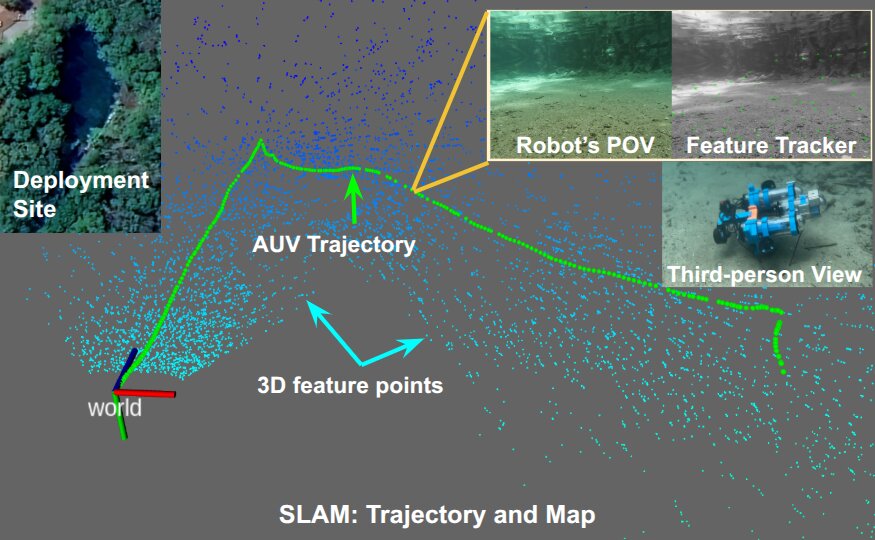}%
     \caption{A rectangle pattern executed by NemeSys is illustrated by the sparse map and AUV trajectory. The trajectory is estimated by a monocular ORB-SLAM3 pipeline. These tests are conducted in a spring water ($1$-$15$\,m depth) environment spanning over a $100$ sq-km area.
     }%
     \vspace{-1mm}
     \label{fig:field_eval}
 \end{figure}

\subsection{Online Exploration in Real World}

Following controlled bench-top validation, \textbf{NemeSys} is deployed in freshwater springs (depth range: $1$-$15$\,m) to assess closed-loop control performance and mission execution under realistic environmental disturbances, including mild currents, variable lighting, and reduced visibility. These field trials are designed to evaluate the system's robustness beyond laboratory conditions, particularly in the presence of flow-induced perturbations and perceptual degradation. For each experiment, a predefined mission script is stored onboard and triggered autonomously during runtime, emulating mid-mission command injection via the ME communication interface without requiring surface intervention.

Fig.~\ref{fig:field_eval} presents a representative rectangular trajectory executed remotely. In the absence of GPS signals, the vehicle trajectory is reconstructed post hoc using ORB-SLAM3~\cite{campos2021orb} applied to forward-facing monocular imagery. Although visual degradation and intermittent feature sparsity occasionally affected SLAM continuity, NemeSys maintained stable depth and heading throughout the mission, demonstrating reliable navigation and control performance in real-world aquatic environments subject to sensing and communication constraints.




\section{Conclusion}
This paper presents {NemeSys}, a novel AUV platform designed to support adaptive autonomy in marine operations. Unlike traditional AUVs that rely on pre-scripted missions, NemeSys introduces a system-level approach to enable low-bandwidth mission updates using ME signals communicated by a remote operator. 
We integrate a stability-aware mechanical configuration with a compact mission encoding framework for structured mission updates. In comprehensive simulation and field trials, NemeSys demonstrates the feasibility of goal-aware replanning and reliable trajectory execution under realistic operating conditions.
Future work will extend this architecture to multi-robot coordination and evaluate long-horizon learning frameworks for predictive mission adaptation and collaborative decision-making.

\bibliographystyle{ieeetr}
\bibliography{robopi_pubs,old_cave_refs,refs}

\end{document}